\documentclass[11pt]{article}

\usepackage[final]{acl}

\usepackage{times}
\usepackage{latexsym}
\usepackage[T1]{fontenc}
\usepackage[utf8]{inputenc}
\usepackage{microtype}
\usepackage{inconsolata}

\usepackage{graphicx}
\usepackage{booktabs}
\usepackage{tabularx}
\usepackage{siunitx}
\usepackage{multirow}
\usepackage{adjustbox}

\usepackage{xeCJK}
\setCJKmonofont{HaranoAjiGothic-Regular.otf}

\usepackage{listings}
\usepackage{xcolor}
\usepackage{upquote}
\usepackage{tikz}
\usetikzlibrary{positioning,arrows.meta}
\usepackage{tabularx}

\newcommand{\best}[1]{\textbf{#1}}

\lstdefinestyle{tightjson}{
  basicstyle=\ttfamily\footnotesize,
  breaklines=true,
  columns=fullflexible,
  keepspaces=true,
  showstringspaces=false,
  frame=single,
  xleftmargin=0pt,
  xrightmargin=0pt,
  aboveskip=0.5em,
  belowskip=0.5em
}

\newenvironment{promptja}
  {\begin{quote}\small}
  {\end{quote}}

\newenvironment{prompten}
  {\begin{quote}\small}
  {\end{quote}}

\lstdefinelanguage{json}{
  morestring=[b]",
  morecomment=[l]{//},
  morekeywords={true,false,null},
  sensitive=false
}

\title{Human-Grounded Multimodal Benchmark with 900K-Scale Aggregated Student Response Distributions from Japan's National Assessment of Academic Ability}

\author{
 \textbf{Kyosuke Takami\textsuperscript{1}},
 \textbf{Yuka Tateisi\textsuperscript{2}},
 \textbf{Satoshi Sekine\textsuperscript{3}},
 \textbf{Yusuke Miyao\textsuperscript{2,3}}
\\
\\
 \textsuperscript{1}Osaka Kyoiku University,
 \textsuperscript{2}University of Tokyo,
 \textsuperscript{3}NII LLMC
\\
 \small{
   \textbf{Correspondence:} \href{mailto:takami-k75@cc.osaka-kyoiku.ac.jp}{takami-k75@cc.osaka-kyoiku.ac.jp}
 }
}

\begin{document}
\maketitle

\begin{abstract}
Authentic school examinations provide a high-validity test bed for evaluating multimodal large language models (MLLMs), yet benchmarks grounded in Japanese K--12 assessments remain scarce. 
We present a multimodal dataset constructed from Japan's National Assessment of Academic Ability, comprising officially released middle-school items in Science, Mathematics, and Japanese Language.
Unlike existing benchmarks based on synthetic or curated data, our dataset preserves real exam layouts, diagrams, and Japanese educational text, together with nationwide aggregated student response distributions (N $\approx$ 900{,}000). 
These features enable direct comparison between human and model performance under a unified evaluation framework. We benchmark recent multimodal LLMs using exact-match accuracy and character-level F1 for open-ended responses, observing substantial variation across subjects and strong sensitivity to visual reasoning demands. Human evaluation and LLM-as-judge analyses further assess the reliability of automatic scoring. Our dataset establishes a reproducible, human-grounded benchmark for multimodal educational reasoning and supports future research on evaluation, feedback generation, and explainable AI in authentic assessment contexts. Our dataset is available at:  \url{https://github.com/KyosukeTakami/gakucho-benchmark}
\end{abstract}

\section{Introduction}

Multimodal large language models (MLLMs) are increasingly evaluated using exam-style benchmarks, yet resources grounded in authentic K–12 assessment contexts remain limited, particularly for Japanese educational settings. Existing benchmarks such as MMLU and C-Eval primarily focus on textual reasoning and only partially capture the multimodal characteristics of real school examinations. In Japanese middle-school assessments, questions often integrate dense visual elements—including diagrams, tables, and visually embedded text—within structured pedagogical layouts designed to assess curriculum-aligned competencies. These authentic design features introduce multimodal reasoning demands that are not fully represented in current evaluation resources, motivating the need for benchmarks derived from real educational assessments. Unlike existing benchmarks, our dataset includes nationwide aggregated student response distributions, enabling direct comparison between human and LLM performance under the same evaluation framework.

\section{Contributions}

\begin{itemize}
\item We introduce a multimodal benchmark derived from Japan’s National Assessment of Academic Ability, preserving authentic exam layouts, diagrams, and Japanese educational text features.
\item The dataset includes nationwide aggregated response distributions (N $\approx$ 900,000), enabling empirical estimation of task difficulty and reasoning patterns.
\item We enable direct comparison between human and LLM performance under a unified evaluation framework, supported by both human annotations and LLM-as-judge scoring.
\end{itemize}

\paragraph{Motivation and challenge.}

\begin{figure}[!htbp]
  \centering
  \includegraphics[width=\linewidth,height=.45\textheight,keepaspectratio]{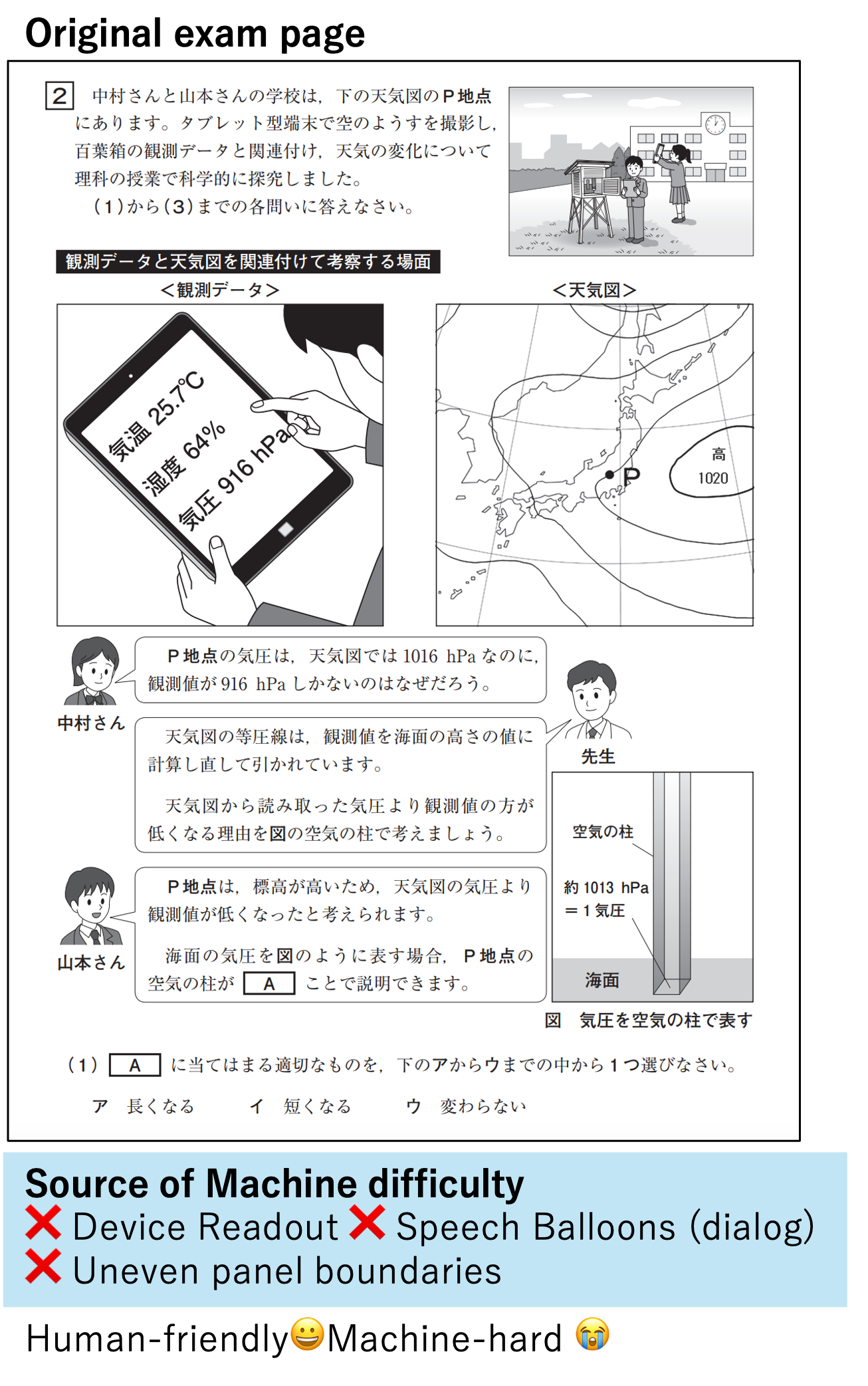}
  \caption{An example of the original exam page.}
  \label{fig:jp-weather-item} 
\end{figure}

Existing benchmarks evaluate LLMs primarily using synthetic or curated datasets, which lack grounding in real-world educational data. In particular, they do not provide population-level response distributions, making it difficult to compare model performance with human learners in a meaningful way.However, constructing such datasets from real-world assessments is itself non-trivial. Raw exam materials are typically released as PDFs with complex multimodal layouts, requiring careful reconstruction of reading order, visual structure, and answer formats before they can be used for systematic evaluation.
As illustrated in Figure~\ref{fig:jp-weather-item} (an example Science item from Japan's National Assessment of Academic Ability), real examination pages combine heterogeneous visual and textual primitives: dialog in speech balloons, numerical readouts on devices, maps with isobars and legends, and both horizontal and vertical Japanese text. Panel boundaries are irregular, reading order is implicit, and labels must be associated with graphical referents. While these multimodal layouts support student comprehension by providing redundancy and context, they introduce bottlenecks for machines—such as OCR errors on vertical text, difficulties in segmenting uneven panels, and challenges in linking callouts to figures.

To address these issues, we propose a pipeline that transforms such human-friendly but machine-unfriendly exam materials into a structured and reproducible multimodal benchmark. Specifically, we segment each page into minimal visual units (e.g., figures, tables, text boxes), assign consistent identifiers and tags, and provide evidence-oriented descriptions that capture quantitative or referential information within each image. We further normalize reading order and cross-panel dependencies through a unified JSON schema that links subquestions to the panels they reference. Each item is then released in dual form—original image(s) and normalized text—to facilitate multimodal reasoning evaluation.

The resulting dataset is derived from Japan’s National Assessment of Academic Ability and includes officially released middle-school questions in Science, Mathematics, and Japanese Language. Each question is represented in a structured format preserving original figures, mathematical expressions, and empirically reported answer distributions. In addition to automatic scoring, human evaluation was conducted for open-ended questions as a preliminary check of scoring validity.

This dataset provides an initial step toward systematic benchmarking of multimodal comprehension in Japanese educational assessment contexts, bridging textual, visual, and cognitive reasoning while establishing reproducible standards for cross-model comparison.

\section{Related Work}

\subsection{Exam-style Benchmarks for LLM Evaluation}

A growing body of work evaluates LLMs using exam-style benchmarks that probe academic knowledge and structured reasoning. 
Widely used datasets such as \textsc{MMLU} \cite{hendrycks2021mmlu}, \textsc{AGIEval} \cite{zong2023agieval}, and \textsc{MMLU-Pro} \cite{zhou2024mmlupro} provide large-scale collections of textual questions across multiple disciplines. 
Language-specific benchmarks including \textsc{C-Eval} \cite{huang2023ceval} and \textsc{CMMLU} \cite{li2023cmmlu}, as well as domain-focused datasets such as \textsc{GPQA} \cite{rein2023gpqa} and mathematical benchmarks like \textsc{GSM8K} \cite{cobbe2021gsm8k} and \textsc{MATH} \cite{hendrycks2021math}, further examine reasoning capabilities.

Recent efforts have incorporated authentic educational materials, including \textsc{RACE} \cite{lai2017race}, \textsc{EXAMS} \cite{hardalov2020exams}, \textsc{GAOKAO-Bench} \cite{zhang2023gaokaobench}, and \textsc{JEEBench} \cite{arora2023jeebench}. 
However, benchmarks reflecting Japanese K--12 assessment practices remain limited, particularly those capturing multimodal reasoning in real exam settings.

\subsection{Multimodal Benchmarks}

Multimodal benchmarks have rapidly expanded to evaluate vision--language understanding. 
Datasets such as \textsc{MS-COCO} \cite{lin2014coco}, \textsc{VQA} \cite{antol2015vqa}, \textsc{GQA} \cite{hudson2019gqa}, and \textsc{NLVR2} \cite{suhr2019nlvr2} assess visual grounding and compositional reasoning, while \textsc{DocVQA} \cite{mathew2021docvqa} and \textsc{TextVQA} \cite{singh2019textvqa} focus on structured visual-text understanding.

Educational multimodal datasets such as \textsc{ScienceQA} \cite{lu2022scienceqa} and \textsc{MMMU} \cite{yue2023mmmu} further introduce diagram-based and textbook-style problems. 
Nevertheless, benchmarks that combine curriculum-aligned tasks with authentic multimodal exam layouts remain scarce, especially in Japanese educational contexts.

\paragraph{Our Position.}
This work bridges exam-style and multimodal evaluation by introducing a curriculum-grounded benchmark derived from Japan’s National Assessment of Academic Ability. 
The dataset captures authentic middle-school examination items combining textual, numerical, and diagrammatic reasoning, and preserves aggregated student response statistics to support reproducible evaluation and human–model comparison.

\section{Dataset Construction}

\begin{figure}[t]
\centering
\begin{tikzpicture}[
    node distance=0.8cm,
    box/.style={
        rectangle,
        draw,
        rounded corners,
        text width=0.82\columnwidth,
        minimum height=1.35cm,
        align=center,
        font=\footnotesize
    },
    arrow/.style={
        -{Latex[length=2mm]},
        thick
    }
]

\node[box] (src) {
\textbf{Official Exam Materials}\\
PDF exam booklets; national assessment reports
};

\node[box, below=of src] (md) {
\textbf{Manual Transcription}\\
Convert to Markdown; preserve reading order; insert image references; use LaTeX for mathematical expressions
};

\node[box, below=of md] (img) {
\textbf{Image Extraction}\\
Extract diagrams and figures; link images to items; preserve visual layout
};

\node[box, below=of img] (json) {
\textbf{Schema-based Conversion}\\
Automatic JSON/JSONL generation; item metadata structuring
};

\node[box, below=of json] (resp) {
\textbf{Response Metadata}\\
Answer distributions; correct / incorrect examples; scoring conditions
};

\node[box, below=of resp] (bench) {
\textbf{Evaluation-ready Benchmark}\\
Multimodal QA tasks; open-ended scoring; LLM evaluation prompts
};

\draw[arrow] (src) -- (md);
\draw[arrow] (md) -- (img);
\draw[arrow] (img) -- (json);
\draw[arrow] (json) -- (resp);
\draw[arrow] (resp) -- (bench);

\end{tikzpicture}
\caption{Pipeline for constructing the benchmark from official exam PDFs and assessment reports.}
\label{fig:pipeline}
\end{figure}
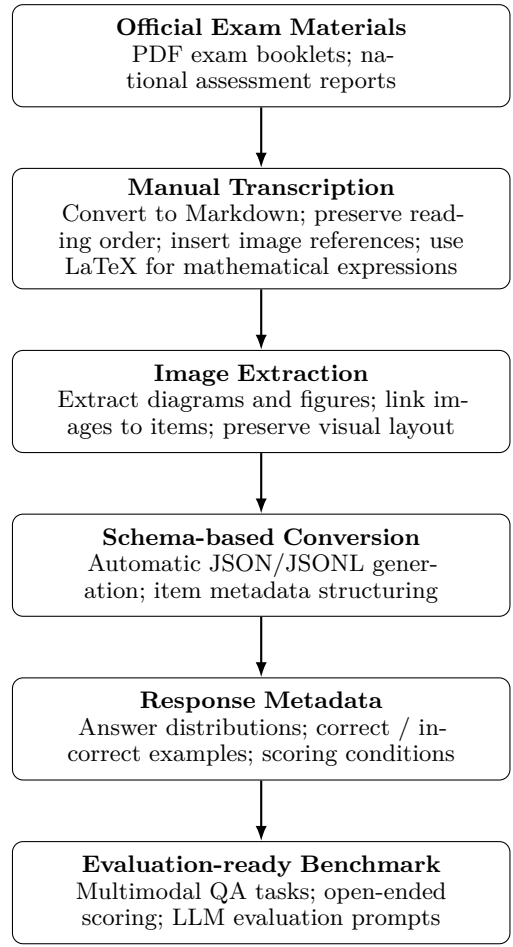

\paragraph{Japan’s National Assessment of Academic Ability.}
Japan’s Ministry of Education, Culture, Sports, Science and Technology (MEXT) has administered the \textit{National Assessment of Academic Ability} annually since FY 2007 for all Grade 6 (elementary) and Grade 9 (middle school) students across the nation, making it a full-population (census-style) survey rather than a sampled one \cite{mext_en_overview_2007,mext_overview_jp}.  
The assessment combines subject tests with student and school questionnaires to support national PDCA-based educational policy and local classroom improvement. The 
core subjects—Japanese and mathematics—are assessed every year, while science and junior-high English are added on a roughly triennial rotation \cite{mext_overview_jp}.  
The National Institute for Educational Policy Research (NIER) publicly releases test items, answer keys, and explanatory materials for each administration \cite{nier_items_2013}, and aggregated datasets are accessible to researchers under a controlled data-lending program with formal usage guidelines \cite{mext_data_lending,edpportal_lending}. Because the dataset is derived from officially
released items of the National Assessment of Academic Ability, the number of items per administration is fixed by the examination design. The current release therefore contains all publicly available Grade 9 items from the 2022 administration.

\begin{table}[!htbp]
\centering
\small
\adjustbox{max width=\linewidth}{
\begin{tabular}{l
    S[table-format=2]
    S[table-format=2]
    S[table-format=2]
    S[table-format=2]
    S[table-format=2.1]}
\toprule
\textbf{Subject} & \textbf{Items} & \textbf{MC} & \textbf{Open} & \textbf{Images} & \textbf{Human Avg. (\%)} \\
\midrule
Science  & 21 & 11 & 10 & 34 & 49.7 \\
Math     & 14 &  4 & 10 & 27 & 52.0 \\
Japanese &  9 &  5 &  4 &  8 & 69.3 \\
\bottomrule
\end{tabular}}
\caption{
Dataset statistics for the 2022 benchmark derived from Japan's National Assessment of Academic Ability.
\textbf{Items} denotes the number of exam questions in each subject;
\textbf{MC} and \textbf{Open} indicate multiple-choice and open-ended items, respectively.
\textbf{Images} counts distinct visual files linked to items and is not additive with item counts.
\textbf{Human Avg. (\%)} indicates the average correct response rate based on nationwide student responses (\(N = 928{,}509\)).
}
\label{tab:year2022-stats}
\end{table}

We obtained the questions and official survey reports on Japanese, mathematics, and science from the 2022 edition of the assessment for Grade 9 students\ref{tab:year2022-stats} to construct JSON data.

The questions
(\cite{kokugo_mondai_2022} for Japanese, 
\cite{suugaku_mondai_2022} for mathematics,
and 
\cite{rika_mondai_2022} for science)
are manually converted to markdown format, and then automatically converted to JSON data. 


\begin{figure}[!htbp]
  \centering
  \includegraphics[width=\linewidth,height=.65\textheight,keepaspectratio]{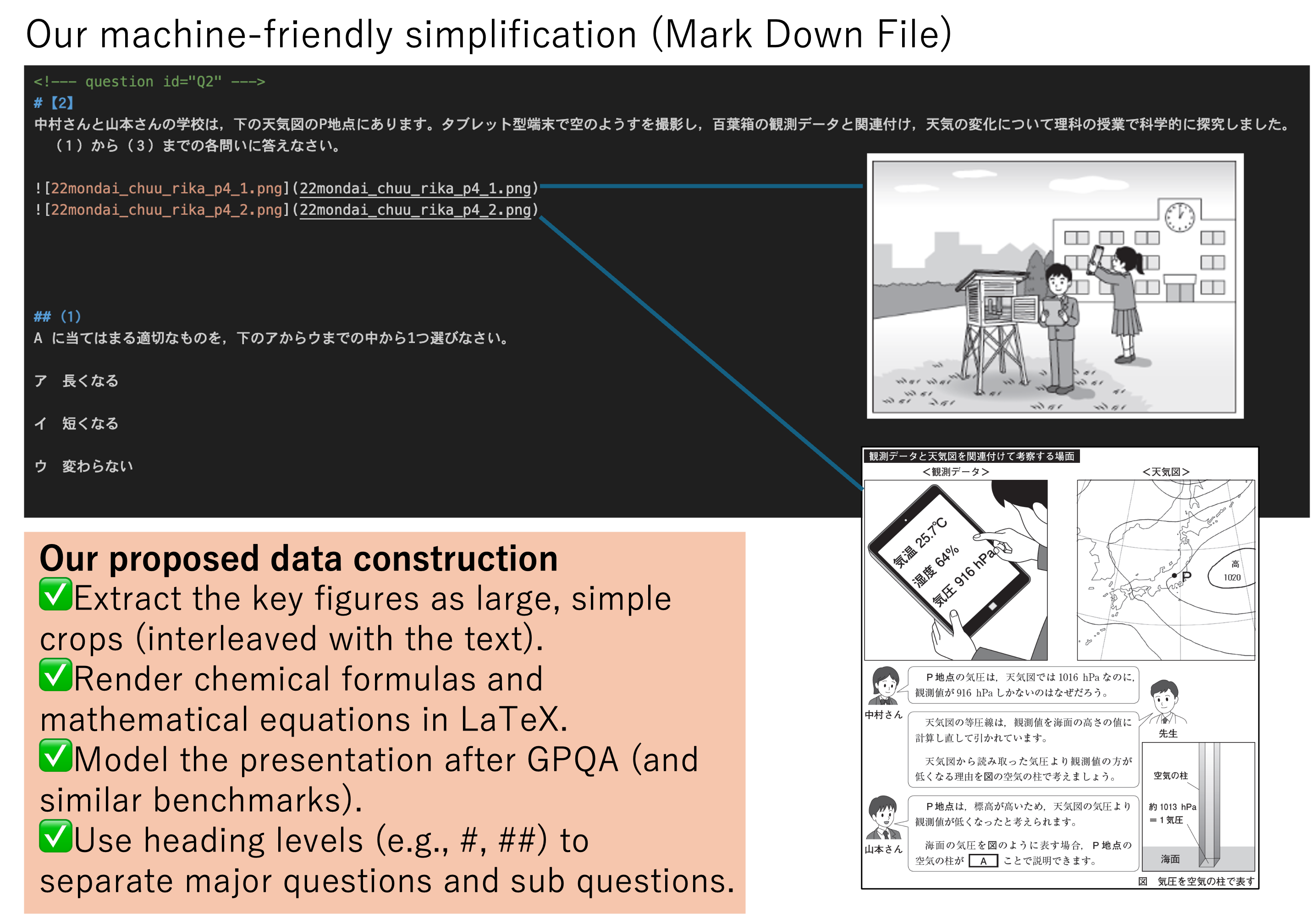}
  \caption{Markdown file for a multimodal question.}
  \label{fig:panel}
\end{figure}

In the Markdown file for the questions, each item is presented as a separate section, with the text reproduced in its original reading order. 
No annotations are added. 
Figures are stored in separate files, and hyperlinks to those files are inserted at the locations where the original illustrations appeared in the text. 
As with the text, no annotations are applied to the figures even when specific locations within a figure may be referenced in the questions. 
Consequently, in the Markdown file as shown in Fig.  \ref{fig:panel}, there is no mechanism to explicitly identify the referenced locations within the figures.
Mathematical expressions appearing in the questions are represented using LaTeX format.

In the official survey report, the percentage of correct responses was aggregated and analyzed across regions and school types nationwide.
The report also provides the intent of the question, a response-type classification table, analytical findings with identified issues, and proposed instructional strategies, for each question. The response-type classification table organizes, in tabular format, the conditions for correct answers, the types of responses observed, the response rates for each type, and whether each type was considered correct.

From the reports of the 2022  assessment for Grade 9 (\cite{kokugo_report_2022} for Japanese, 
\cite{suugaku_report_2022} for mathematics, and
\cite{rika_report_2022}), 
we extracted the response-type classification tables into a standardized JSON format for subsequent analysis. When exemplar responses corresponding to each type were available—either within the classification tables or embedded in the analytical commentary—they are also included in the JSON data.

\subsection{Example JSON Item and Its Educational Significance}

We illustrate the structure of our dataset using an example item from the \textit{National Assessment of Academic Ability (2022, Middle School Mathematics)}. 
The dataset preserves key features of Japanese educational materials, including ruby annotations, mathematical expressions, and references to diagrams.

Each record additionally includes large-scale, empirically grounded response data. 
The \texttt{answer\_distribution} field encodes response proportions from approximately \textbf{900{,}000 middle-school students} nationwide, categorized into fine-grained reasoning types. 
The dataset also provides representative \texttt{correct\_examples} and \texttt{incorrect\_examples}, capturing diverse valid and invalid reasoning patterns.

These components enable direct comparison between human and LLM performance and support realistic modeling of reasoning diversity. 
As a result, the dataset provides a foundation for reproducible evaluation and future work on formative feedback and explainable assessment design.

A full JSON example is provided in Appendix~A.

\section{Evaluation}

We evaluate a set of representative large multimodal language models on the proposed National Assessment benchmark in order to provide baseline performance references.
The selected models include widely used API-accessible systems as well as a small number of open-weight models that can be executed locally.

For clarity and reproducibility, Table~\ref{tab:model-list} lists the canonical names used throughout this paper together with their providers and exact API identifiers or model repository names.
These models were chosen because they have demonstrated strong performance on recent vision--language reasoning and multimodal understanding tasks.

Our evaluation is intended to validate the usefulness of the dataset as a benchmark rather than to provide an exhaustive comparison of model architectures.
Future work will extend the evaluation to additional open-weight models and emerging multimodal systems.

\begin{table}[t]
\centering
\footnotesize
\adjustbox{max width=\linewidth}{
\begin{tabular}{l l l}
\toprule
\textbf{Canonical name} & \textbf{Provider} & \textbf{API identifier} \\
\midrule
GPT-5               & OpenAI     & \texttt{gpt-5} \\
GPT-4o              & OpenAI     & \texttt{gpt-4o} \\
GPT-4o-mini         & OpenAI     & \texttt{gpt-4o-mini} \\
Gemini 3.1 Pro      & Google     & \texttt{gemini-3.1-pro-preview} \\
Gemini 2.5 Pro      & Google     & \texttt{gemini-2.5-pro} \\
Gemini 2.5 Flash    & Google     & \texttt{gemini-2.5-flash} \\
Claude Sonnet 4.5   & Anthropic  & \texttt{claude-sonnet-4-5-20250929} \\
Claude Sonnet 4     & Anthropic  & \texttt{claude-sonnet-4-20250514} \\
Claude 3.7 Sonnet   & Anthropic  & \texttt{claude-3-7-sonnet-20250219} \\
\midrule
\multicolumn{3}{l}{\textit{Open-weight models}} \\
Qwen2.5-VL-7B-Instruct & Alibaba & \texttt{Qwen/Qwen2.5-VL-7B-Instruct} \\
Gemma 3 12B-it         & Google   & \texttt{gemma-3-12b-it} \\
\bottomrule
\end{tabular}}
\caption{Models in the main multimodal evaluation. Canonical names used throughout the paper map to the exact API identifiers shown here.}
\label{tab:model-list}
\end{table}

\section{Prompting}

We use two prompt types aligned with task format: (i) label-only output for multiple-choice items and (ii) answer-only output for open-ended items. Prompts are written in Japanese to match the source data.

For reproducibility, full prompt templates (Japanese and English) are provided in Appendix~B. And more detailed information about evaluation methods are in Appendix~C,D and E.

\section{Experimental Results}

\begin{table*}[ht]
\centering
\scriptsize
\setlength{\tabcolsep}{4pt}
\begin{adjustbox}{max width=\textwidth}
\begin{tabular}{@{}l S S S S S S@{}}
\toprule
\multirow{2}{*}{\textbf{Model}} &
\multicolumn{2}{c}{\textbf{Science} (n=21; MC=11, Open=10)} &
\multicolumn{2}{c}{\textbf{Math} (n=14; MC=4, Open=10)} &
\multicolumn{2}{c}{\textbf{Japanese} (n=9; MC=5, Open=4)} \\
\cmidrule(lr){2-3}\cmidrule(lr){4-5}\cmidrule(lr){6-7}
 & {\textbf{MC}} & {\textbf{F1}} & {\textbf{MC}} & {\textbf{F1}} & {\textbf{MC}} & {\textbf{F1}} \\
\midrule
GPT-5              & \best{0.909} & 0.489 & 0.750 & 0.623 & 0.800 & 0.324 \\
GPT-4o             & 0.455 & 0.399 & \best{1.000} & 0.492 & 0.600 & 0.278 \\
GPT-4o-mini        & 0.455 & 0.379 & 0.750 & 0.461 & 0.800 & 0.242 \\
Gemini 3.1 Pro     & 0.636 & 0.324 & 1.000 & 0.358 & 0.800 & \best{0.519} \\
Gemini 2.5 Pro     & 0.727 & 0.498 & 0.750 & 0.558 & 0.800 & 0.250 \\
Gemini 2.5 Flash        & 0.727 & \best{0.661} & 0.750 & 0.542 & 1.000 & 0.250 \\
Claude Sonnet 4.5  & 0.636 & 0.521 & \best{1.000} & \best{0.695} & \best{1.000} & 0.328 \\
Claude Sonnet 4    & 0.727 & 0.548 & 0.750 & 0.548 & 0.600 & 0.342 \\
Claude 3.7 Sonnet  & 0.545 & 0.447 & 0.750 & 0.663 & \best{1.000} & 0.311 \\
\midrule
\multicolumn{7}{l}{\textit{Open-weight multimodal models}} \\
Qwen2.5-VL-7B-Instruct & {0.363} & {0.349} & {0.750} & {0.195} & {0.400} & {0.305} \\
Gemma-3-12B-it         & {0.545} & {0.329} & {0.250} & {0.246} & {0.400} & {0.268} \\
\bottomrule
\end{tabular}
\end{adjustbox}
\caption{Summary across three subjects with multiple-choice (MC) accuracy and character-level F1 for open-ended responses. Bold indicates the best value within each subject and metric. Open-weight models are evaluated to provide indicative local baselines.}
\label{tab:all-subjects-updated}
\end{table*}

\paragraph{Overall analysis across subjects.}
Table \ref{tab:all-subjects-updated} shows overall analysis across subjects indicating EM accuracy for MC questions and F1 score for Open questions. Across Science, Math, and Japanese, the results reveal that multimodal large language models (MLLMs) exhibit distinct subject-specific behaviors, yet a consistent benefit from visual information. In Science, models such as \textbf{GPT-5} and \textbf{Gemini 2.5} demonstrate strong performance. Math shows comparatively stable multiple-choice accuracy across models (around 0.85 on average), but larger variation in open-ended F1. In Japanese Language, performance diverges—\textbf{Gemini 2.5} achieves the best multiple-choice accuracy (0.800), whereas \textbf{Claude Sonnet 4} leads on open-ended generation (0.342). 

\paragraph{Do images matter?}
We conduct an ablation comparing performance with and without images. Results (Appendix~F) show that images substantially improve performance in Science, while effects are mixed in Math and Japanese, reflecting task-dependent reliance on visual information.

\section{Human Evaluation vs LLM-as-a-judge}

\paragraph{Human Evaluation}
Two of the authors independently conducted the human evaluation. 
The evaluation focused exclusively on \textit{open-ended} questions; multiple-choice (MC) items were automatically scored and thus excluded from manual assessment. 
Each rater evaluated the correctness of the LLM outputs by referencing the information defined in the dataset schema—specifically the fields \texttt{correct\_answer}, \texttt{correct\_condition}, \texttt{answer\_distribution}, \texttt{correct\_examples}, and \texttt{incorrect\_examples}—as described in the Example JSON. 
The raters judged whether each generated response satisfied the prescribed correct condition and corresponded to the officially correct answer type. 
One of the raters, who has professional experience as a K–12 teacher, supervised the grading process and ensured that the evaluation rubric and scoring decisions adhered to pedagogically sound practices consistent with classroom-based assessment. 
This procedure provided a preliminary check of scoring consistency and helped align the evaluation protocol with the educational intent of the dataset. However, because the evaluation was conducted by two authors rather than independent external annotators, we treat these results as an initial validation rather than definitive evidence of scoring reliability.

\paragraph{LLM-as-a-judge.}

In our work, we also adopt the paradigm of using a large language model (LLM) as an automated evaluator of other LLM outputs—commonly referred to as “LLM-as-a-judge.” This approach has gained traction because models such as GPT-4o can achieve strong agreement with human judgments in open-ended, generative settings \cite{nichols2023judging,huang2024llmjudge}. By deploying GPT-4o to score responses to open-ended items in our dataset, we gain scalable, consistent evaluation across multiple models and subject areas. Biases inherent to LLM-based judges (such as self-preference, prompt-sensitivity, or domain misalignment) may distort comparative outcomes and require future calibration or hybrid human-machine assessment strategies \cite{dorner2025limits,huang2024llmjudge}.

\begin{table*}[t]
\centering
\scriptsize
\setlength{\tabcolsep}{3pt}
\renewcommand{\arraystretch}{0.92}
\begin{adjustbox}{max width=\textwidth}
\begin{tabular}{lrrrr}
\toprule
\textbf{Model} & \textbf{Human Evaluator Acc. (2 humans, \%)} & \textbf{LLM-as-a-judge Acc. (\%)} & \textbf{Agree (2 humans, \%)} & \textbf{Cohen's $\kappa$ (2 humans)} \\
\midrule
\multicolumn{5}{l}{\textbf{Science (Open=10, Students' ACC. 57.7) }}\\
GPT-5         & 75.0 & 50.0 & 90.0 & 0.74 \\
GPT-4o         & 55.0 & 40.0 & 70.0 & 0.44 \\
GPT-4o-mini    & 30.0 & 20.0 & 80.0 & 0.52 \\
Gemini 3.1 Pro& 70.0 & 60.0 & 80.0 & 0.55 \\
Gemini 2.5 pro & 70.0 & 70.0 & 100.0 & 1.00 \\
\textbf{gemini 2.5 Flash} & \textbf{90.0} & \textbf{80.0} & \textbf{100.0} & \textbf{1.00} \\
Claude Sonnet 4.5     & 45.0 & 30.0 & 90.0 & 0.80 \\
Claude Sonnet 4       & 60.0 & 50.0 & 100.0 & 1.00 \\
Claude Sonnet 3.7     & 65.0 & 50.0 & 90.0 & 0.78 \\
\cmidrule(lr){1-5}
\multicolumn{5}{l}{\textit{Open-weight multimodal models}}\\
Qwen2.5-VL-7B & 30.0 & 20.0 & 100.0 & 1.00 \\
Gemma-3-12B-it   & 30.0 & 40.0 & 100.0 & 1.00 \\
\midrule
\multicolumn{5}{l}{\textbf{Math (Open=10, Students'Acc. 53.1)}}\\
GPT-5         & 85.0 & 60.0 & 70.0 & 0.00 \\
GPT-4o         & 55.0 & 30.0 & 70.0 & 0.40 \\
GPT-4o-mini    & 45.0 & 30.0 & 90.0 & 0.80 \\
Gemini 3.1 Pro & 55.0 & 80.0 & 90.0 & 0.80  \\
Gemini 2.5 Pro & 80.0 & 60.0 & 80.0 & 0.41 \\
\textbf{Gemini-2.5flash} & \textbf{90.0} & 60.0 & \textbf{100.0} & \textbf{1.00} \\
\textbf{Claude Sonnet 4.5} & \textbf{90.0} & \textbf{80.0} & \textbf{100.0} & \textbf{1.00} \\
Claude Sonnet 4       & 75.0 & 80.0 & 90.0 & 0.74 \\
Claude Sonnet 3.7     & 85.0 & 80.0 & 90.0 & 0.62 \\
\cmidrule(lr){1-5}
\multicolumn{5}{l}{\textit{Open-weight multimodal models}}\\
Qwen2.5-VL-7B & 30.0 & 40.0 & 100.0 & 1.00 \\
Gemma-3-12B-it   & 45.0 & 60.0 & 90.0 & 0.80 \\
\midrule
\multicolumn{5}{l}{\textbf{Japanese (Open=4, Students'Acc. 71.2)}}\\
\textbf{GPT-5} & \textbf{37.5} & 0.0 & 75.0 & 0.50 \\
GPT-4o         & 25.0 & 0.0 & \textbf{100.0} & \textbf{1.00} \\
GPT-4o-mini    & 25.0 & 0.0 & 50.0 & 0.00 \\
Gemini 3.1 Pro & \textbf{37.5} & 0.0 & 75.0 & 0.50 \\
Gemini 2.5 Pro & 25.0 & 25.0 & \textbf{100.0} & \textbf{1.00} \\
Gemini 2.5 Flash    & 25.0 & 25.0 & \textbf{100.0} & \textbf{1.00} \\
Claude Sonnet 4.5     & 25.0 & 25.0 & \textbf{100.0} & \textbf{1.00} \\
Claude Sonnet 4       & 25.0 & 25.0 & \textbf{100.0} & \textbf{1.00} \\
Claude Sonnet 3.7     & 25.0 & 25.0 & \textbf{100.0} & \textbf{1.00} \\
\cmidrule(lr){1-5}
\multicolumn{5}{l}{\textit{Open-weight multimodal models}}\\
Qwen2.5-VL-7b & 25.0 & 0.0 & 50.0 & 0.00  \\
Gemma-3-12B-it   & 0.0 & 0.0 & 100.0 & N/A (single-class annotations) \\
\bottomrule
\end{tabular}
\end{adjustbox}
\caption{Open-ended questions: human evaluation vs.\ LLM as a judge (GPT-4o). “Human Evaluator Acc.” is the two-rater average accuracy; “LLM-Judge Acc.” is automatic scoring accuracy on the same items. Open-weight models are included to provide indicative reproducible baselines.}
\label{tab:open-human-vs-judge}
\vspace{-2mm}
\end{table*}


\subsection{Interpreting Human vs.\ LLM-as-a-judge Results}

Table~\ref{tab:open-human-vs-judge} contrasts two-rater human judgments with GPT-4o as an automatic judge
for open-ended items. Human scoring showed relatively high raw agreement in many
conditions, although the small number of open-ended items, especially in Japanese,
makes reliability estimates unstable. Therefore, these results should be interpreted
as preliminary evidence of scoring consistency rather than as conclusive validation
of the rubric. The lone anomaly (\textit{Math/gpt-5}: 70\% agreement,
$\kappa = 0.00$) is consistent with a prevalence effect, where class imbalance can
depress Cohen's $\kappa$ despite moderate raw agreement.

Table~\ref{tab:human-judge-corr} quantifies alignment between human accuracy and GPT-4o judge accuracy across models. Correlation is \textbf{strong in Science} (all models: $r{=}0.899$; w/o open-weight: $r{=}0.934$) and \textbf{moderate in Math} (all models: $r{=}0.600$; w/o open-weight: $r{=}0.582$), indicating that GPT-4o largely preserves model \emph{rank ordering}, with greater sensitivity in Math. By contrast, for \textbf{Japanese} the correlation is near zero across all models ($r{\approx}0$) and negative without open-weight models ($r=-0.598$). This likely reflects (i) a restricted dynamic range with human accuracies near 25\%, (ii) a small sample size (four open-ended items), and (iii) skew from cases where the judge assigns 0\%. These factors weaken both rank stability and comparability, suggesting that LLM-as-judge for Japanese should be used with caution, with additional human verification and prompt/rubric calibration.

\begin{table}[t]
\centering
\small
\begin{tabular}{lcc}
\toprule
\textbf{Subject} & \textbf{All Models} & \textbf{w/o Open-weight} \\
\midrule
Science   & 0.899 & \textbf{0.934} \\
Math      & 0.600 & 0.582 \\
Japanese  & $\approx 0^{\dagger}$ & -0.598 \\
\bottomrule
\end{tabular}
\caption{Correlation between human and LLM-judge accuracy.
$^{\dagger}$ The computed value is approximately $1.4 \times 10^{-17}$, 
but is not meaningful due to limited variance and small sample size.}
\label{tab:human-judge-corr}
\end{table}

\section{Error Analysis comparison with students answer rate}

\begin{figure}[h]
\centering
\includegraphics[width=0.95\columnwidth]{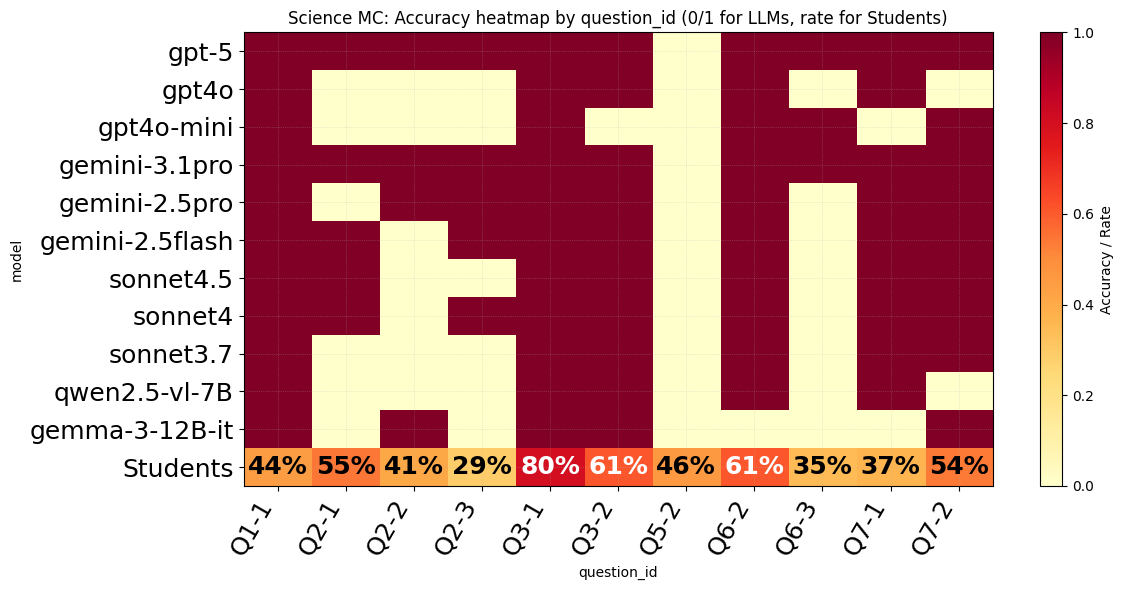}
\includegraphics[width=0.95\columnwidth]{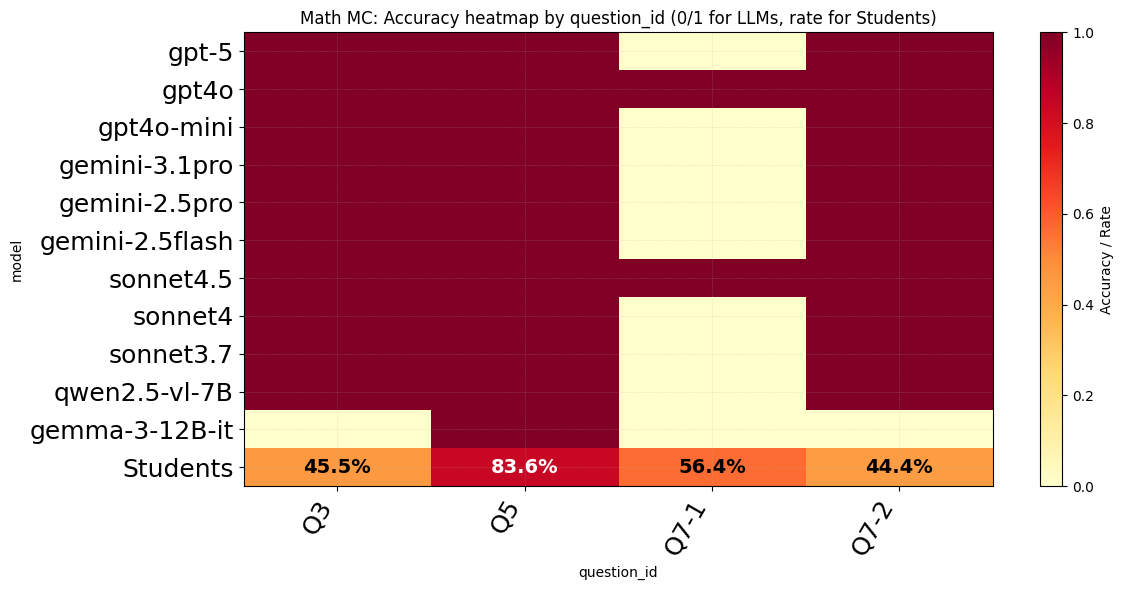}
\includegraphics[width=0.95\columnwidth]{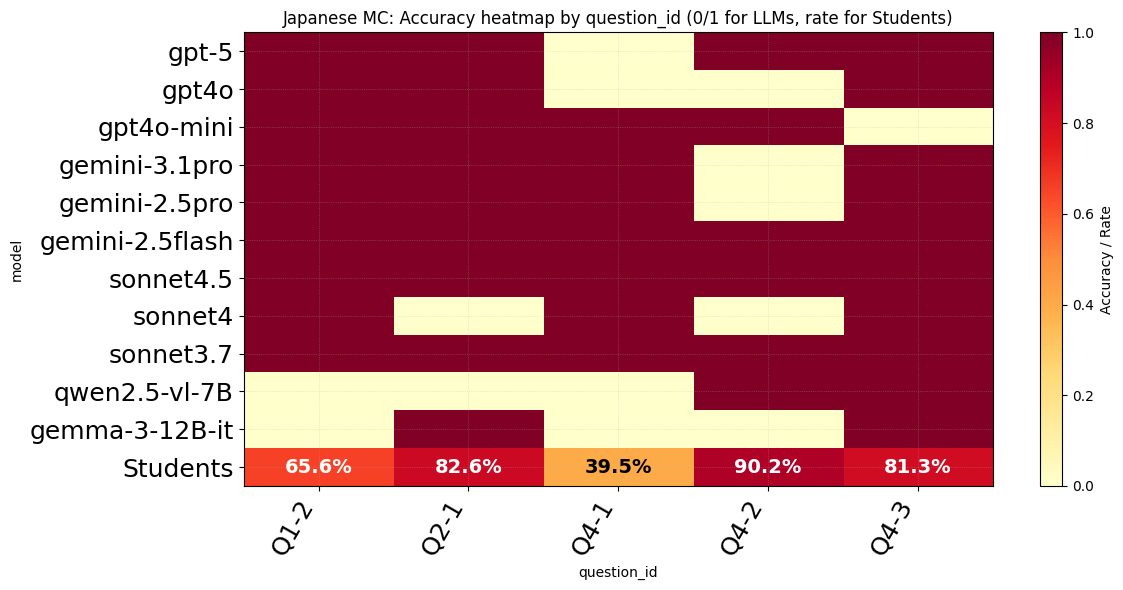}
\caption{
Accuracy heatmaps for multiple-choice (MC) items across three subjects.
Each row represents a model, and each column corresponds to an item ID.
Color intensity encodes model accuracy (0 or 1) for LLMs and empirical correct-answer rates for students.
Lighter cells indicate correctly answered items, while darker ones indicate incorrect predictions.
Student answer rates (bottom row) are annotated for comparison between human and model performance.
}
\label{fig:mc-heatmaps}
\end{figure}

\paragraph{Visualization of multiple-choice performance.}
Figure~\ref{fig:mc-heatmaps} visualizes item-level multiple-choice (MC) accuracy as heatmaps for \textbf{Science}, \textbf{Math}, and \textbf{Japanese}. 
Each matrix row represents a large language model (LLM), and each column corresponds to a question ID within the subject. 
Color intensity encodes model correctness (binary 0/1) for LLMs and student answer rates (in percent) for the bottom row. 
This visualization allows direct comparison of model accuracy patterns against human performance distributions.
In \textbf{Science}, accuracy varies substantially across items, with several models performing well on questions. Notably, on problems that humans often get wrong but that LLMs alternately answer correctly or incorrectly (Fig. \ref{fig:mc-heatmaps} top: Q2-3, Q6-3, Q7-1) differences in visual skill become salient—diagram reading, cross-panel integration, and scale interpretation. This pattern suggests current MLLMs lean on linguistic or pattern-based cues and do not reliably reproduce the visual-measurement strategies human test-takers deploy.
In \textbf{Math}, LLMs generally achieve high consistency on simple numerical problems but exhibit variability on those involving questions that require interpreting diagrams and graphs in images. For example, items such as Q7-1 in Fig \ref{fig:mc-heatmaps}. middle —where LLMs exhibit high error rates—are effectively unsolvable without reading the accompanying histogram; correct responses hinge on explicit visual parsing rather than textual cues.
In contrast, \textbf{Japanese} items show relatively lower model performance overall, likely due to challenges in interpreting language embedded in images and vertically formatted text. In particular as shown in Fig. \ref{fig:mc-heatmaps} bottom, Q4-2—on which LLMs performed especially poorly—requires examinees to inspect an image of a kanji character and explain why its visual balance is deficient. The item is effectively unanswerable without reliable recognition of stroke count and intra-character spatial proportioning (e.g., alignment and symmetry), a fine-grained visual analysis that current models often fail to perform. Moreover, on the open-ended tasks the models uniformly failed on kanji dictation (kakitori), unable to retrieve or compose the required characters. We also observed cases where prompts explicitly required “quote the sentence from the image,” yet the models produced fabricated, plausible-sounding text rather than a verbatim quotation. Taken together, these errors indicate deficits in orthographic recall, precise image-text grounding (OCR fidelity), and adherence to source when exact transcription is required.

\section{Discussion and Future work}

This work shows that a single-year slice of Japan’s National Assessment of Academic Ability can be turned into a legally clean, curriculum-grounded multimodal benchmark with reproducible scoring. Yet restricting the release to one administration limits statistical power and construct coverage (e.g., formats that rotate across years). Because the assessment is nationwide and items are released annually, the natural next step is a longitudinal scale-up: a multi-year benchmark enabling (i) year-over-year tracking under natural distribution shift (topic mix, layout, vertical text density), (ii) stronger reliability via repeated constructs, and (iii) temporally stratified evaluation to mitigate contamination by holding out post-training administrations.

A second direction is to leverage the dataset’s empirically grounded response-type distributions (from cohorts of about 900k students) for \emph{population-aware} evaluation. Beyond point accuracy, future releases can report distributional distances between model outputs and human response profiles at the item level, exposing whether model errors align with real student misconceptions. With aggregated response tables (as in this paper) and, where permitted, microdata via the official lending program, one can calibrate item difficulty proxies and, in the longer term, fit psychometric models (e.g., Rasch/2PL; cognitive diagnostic formulations) to support simulation studies for learning diagnostics and policy analysis \cite{embretson2000item,baker2004basics,rupp2010diagnostic}. Such simulations would allow researchers to estimate how changes in item pools or instruction affect cohort-level mastery patterns before classroom deployment.

The presence of \texttt{correct\_examples}, \texttt{incorrect\_examples}, and fine-grained \texttt{answer\_distribution} also opens a third avenue: \emph{grounded synthetic data}. Prior work shows that synthetic examples can improve robustness when guided by high-quality seeds and task-aware constraints \cite{feng2023survey,long2024llm}. Here, seeds anchored in authentic student behaviors (valid solutions and common errors with observed frequencies) enable (i) rubric-faithful augmentation for training graders or feedback generators, (ii) distractor generation reflecting real misconceptions, and (iii) counterfactual tests targeting specific reasoning skills. Because our schema preserves multimodal evidence links, future augmentation can extend beyond text to vision-grounded references (e.g., scales, callouts, panel IDs), improving ecological validity.

\section*{Limitations}

This work introduces an initial multimodal benchmark linked to large-scale aggregated student response distributions; however, several limitations should be noted.

\textbf{First, domain specificity.}
The dataset is based on a single national assessment from Japan, reflecting specific curricular and linguistic characteristics. This may limit generalizability to other educational contexts.

\textbf{Second, temporal coverage.}
The current benchmark is constructed from a single year of assessment data. While we are actively expanding the dataset to include multiple years, the present version does not capture longitudinal variations in item design or student performance.

\textbf{Third, preprocessing and modality constraints.}
Although multimodal information (e.g., diagrams and layouts) is preserved, preprocessing steps such as segmentation and normalization may introduce noise or omit fine-grained visual details.

\textbf{Fourth, aggregated response data.}
The benchmark uses large-scale aggregated student responses rather than individual-level interaction data, limiting analysis of learning processes and reasoning behaviors.

\textbf{Fifth, evaluation limitations.}
Automatic metrics for open-ended responses may not fully capture semantic correctness or diverse valid answers.

Future work will address these limitations by expanding to multi-year datasets, improving multimodal fidelity, and incorporating richer evaluation and process-level data.

We also note that benchmarks derived from standardized educational assessments may reflect institutional and curricular biases, which should be considered when interpreting model performance.

\section*{Acknowledgments}
This work was supported by JSPS KAKENHI Grant Numbers JP23K17012, JP23K25698.

\bibliography{custom}

\appendix


\section{Example JSON Item}

\begin{lstlisting}[style=tightjson,caption={Example JSON object for an open-ended mathematics question (bilingual description).},label={lst:json-math-example}]
{
  "source": "National Assessment of Academic Ability",
  "subject": "Middle School Mathematics",
  "year": "2022",
  "question_id": "Q6-2",
  "label": "問6（2）",
  "main_text": "康太さんは，2つの偶数の和がどのような場合に4の倍数になるかを調べています。 ...",
  "sub_text": "（２）康太さんは，$2＋6＝8$のように，同じ2つの偶数の和のほかにも ...",
  "main_image_files": ["22mondai_chuu_suugaku_p9_1.png"],
  "sub_image_files": ["22mondai_chuu_suugaku_p11_1.png","22mondai_chuu_suugaku_p11_2.png"],
  "answer_style": "openEnded",
  "correct_answer": "$4(n+1)$\n$n+1$は整数だから、$4(n+1)$は４の倍数である。\nしたがって、差が４である２つの偶数の和は、４の倍数になる。",
  "correct_condition": "＜$4(n + 1)$ と計算している場合＞ (a) $n + 1$ は整数だから、(b) $4(n + 1)$ は４の倍数である。",
  "answer_distribution": [
    {"type_id":1,"answer_type":"$4(n+1)$の場合：(a)、(b)について記述しているもの。","response_rate_percent":20.0,"correct":true},
    {"type_id":6,"answer_type":"$4n+4$の場合：(c)、(d)について記述しているもの。","response_rate_percent":2.0,"correct":true},
    {"type_id":11,"answer_type":"４×□ の□に$(n+1)$以外の文字を用いたもの。","response_rate_percent":3.9,"correct":false},
    {"type_id":99,"answer_type":"上記以外の解答","response_rate_percent":19.4,"correct":false},
    {"type_id":0,"answer_type":"無解答","response_rate_percent":19.6,"correct":false}
  ],
  "correct_examples": [
    "$4(n+1)$\n$n+1$は整数だから、$4(n+1)$は４の倍数である。",
    "$4n+4$\n$4n$、４が４の倍数であるから、$4n+4$は４の倍数である。"
  ],
  "incorrect_examples": [
    "$2n+(2n+4)=2n+6n=8n$",
    "$2n+(2n+4)=4n^2+8n$"
  ]
}
\end{lstlisting}

\section{Prompting}
\paragraph{User prompt: MC (label-only).}
We concatenate the main stem and sub-item text and require a \emph{single label character} (e.g., ア/イ/A/1) on one line, with no explanation. We do not append the full option texts to the prompt by default, which reduces verbosity and encourages grounding in the visual context.

\noindent\textit{Japanese (original):}
\begin{promptja}
次の問題に答えてください。回答は **選択肢のラベル1文字のみ**（例：ア, イ, A, 1 など）で返してください。理由や説明は禁止です。
出力フォーマット：ラベルのみ（例：ア）
\end{promptja}

\noindent\textit{English (faithful gloss):}
\begin{prompten}
Answer the following question. Return \textbf{exactly one label character} (e.g., ア, イ, A, 1) with no explanation.
Output format: label only (e.g., ア)
\end{prompten}

\paragraph{User prompt: Open-ended (answer-only).}
We similarly provide the stem/sub-item text but require a \emph{one-line answer} only, with no rationale or extra tokens; blank outputs are disallowed.

\noindent\textit{Japanese (original):}
\begin{promptja}
次の問題に答えてください。

出力フォーマット：回答のみ（余計な説明は書かない）
\end{promptja}

\noindent\textit{English (faithful gloss):}
\begin{prompten}
Answer the following question.

Output format: answer only (do not add any extra explanation)
\end{prompten}

\paragraph{Model-specific tightening (for \texttt{gpt-5}).}
In pilot runs we observed occasional \emph{silent} completions (empty strings) or overlong answers from \texttt{gpt-5} on label-only and answer-only prompts. To suppress both failure modes, we append a \emph{strict, one-line constraint} only when the target model name contains \texttt{gpt-5}, explicitly banning extraneous symbols and rationales. This mirrors best practices for output reliability—constraining the format and, if needed, retrying with bounded backoff on transient errors—while keeping the scoring contract unchanged. We also note that OpenAI’s GPT-5 documentation emphasizes strengthened steerability and developer controls \citep{openai_gpt5_intro,openai_gpt5_prompting}, which motivates the lightweight, model-targeted tightening rather than dataset-wide prompt changes.

\noindent\textit{Japanese (original):}
\begin{promptja}
【厳守】上記の指示どおり、1行で出力。指定以外の記号・説明は禁止。空欄は禁止。
\end{promptja}

\noindent\textit{English (faithful gloss):}
\begin{prompten}
[STRICT] Follow the above instructions and output on a single line. Do not include any symbols or explanations beyond the specification. Blank output is forbidden.
\end{prompten}

\section{Image handling and routing.}

For items that include images, we pass \emph{all associated images}—whether a single figure or multiple panels—into the same prompt turn in normalized reading order. To ensure fairness across providers and comply with size limits, images are pre-encoded as data URLs and, when necessary, automatically \emph{resized and recompressed} (preserving aspect ratio and legibility) so that each file stays under a unified per-image cap of 7 MB and the total inline payload remains within $\sim$20 MB. Each record is routed to MC or open-ended mode automatically; predictions are then normalized (MC label mapping) or scored with exact match and character-level F1 (open-ended). The released JSONL logs per-item \texttt{accuracy} (MC) and \texttt{f1\_chara} (open) together with raw predictions.

\section{Evaluation Methodology.}
We adopt an evaluation methodology adapted from the MMLU framework, following standard practices in large language model evaluation such as EleutherAI’s \texttt{lm-harness} and OpenAI Evals.
We implemented a unified evaluation pipeline, which normalizes input processing across providers (\texttt{OpenAI}, \texttt{Gemini}, and \texttt{Claude}) and enforces a per–image cap of 7\,MB for fairness.
Each question record is formatted using a strict Japanese instruction template, and models are prompted to output only the final answer (a single label or concise free–form response) without explanations.
The client automatically resizes large figures to comply with token and bandwidth limits while preserving the original image information.
All model outputs, together with metadata (question ID, task type, prompt text, image file paths, and labels), are saved as JSONL for reproducibility and further statistical analysis.

\section{Evaluation Metrics.}
Our primary metric for multiple–choice items is \textbf{accuracy}, computed as the proportion of correctly predicted labels.
For open–ended items, we report both \textbf{Exact Match (EM)} and character–level \textbf{F1}, following conventions in Japanese NLP and reading comprehension evaluation.
The F1 score is computed by comparing the predicted and gold text at the character level, providing robustness to minor lexical or inflectional variations.
In addition, per–question results are recorded, and a summary JSONL is exported containing both \texttt{accuracy} and \texttt{f1\_chara} fields appended to the original gold dataset.
This unified scheme enables direct comparison across models and providers under consistent multimodal input conditions, ensuring that accuracy and F1 reflect both discrete answer correctness and generative quality.

\begin{table}[t]
\centering
\small
\adjustbox{max width=\linewidth}{
\begin{tabular}{l
                S[table-format=1.3] S[table-format=1.3] S[table-format=+1.3]
                S[table-format=1.3] S[table-format=1.3] S[table-format=+1.3]}
\toprule
& \multicolumn{3}{c}{Multiple-choice Accuracy} &
  \multicolumn{3}{c}{Open-ended F1 (char)} \\
\cmidrule(lr){2-4}\cmidrule(lr){5-7}
Subject & {w/ img} & {w/o img} & {$\Delta$} &
          {w/ img} & {w/o img} & {$\Delta$} \\
\midrule
Science  & 0.646 & 0.535 & +0.111 & 0.474 & 0.354 & +0.120 \\
Math     & 0.833 & 0.778 & +0.056 & 0.549 & 0.552 & -0.003 \\
Japanese & 0.822 & 0.867 & -0.044 & 0.316 & 0.310 & +0.006 \\
\midrule
All (weighted) & 0.728 & 0.667 & +0.061 & 0.479 & 0.429 & +0.050 \\
\bottomrule
\end{tabular}}
\caption{Average performance with vs. without images across the nine-model set, excluding open-weight multimodal models. The no-image condition removes visual inputs while preserving the textual prompt.}
\label{tab:with-vs-without-images-revised}
\end{table}

\section{Impact of Visual Inputs}
Table~\ref{tab:with-vs-without-images-revised} reports the updated average performance across the revised nine MLLMs with and without images. Providing images substantially improves \textbf{Science} performance (MC $+0.111$, F1$_\text{char}$ $+0.120$), confirming that our items require grounding textual responses in visual information such as diagrams, graphs, and experimental setups. For \textbf{Math}, the MC gain is moderate ($+0.056$), while F1 remains nearly unchanged ($-0.003$), suggesting that visual input mainly helps with answer selection rather than materially improving free-form response quality. In \textbf{Japanese}, the pattern is mixed: MC slightly declines ($-0.044$), whereas F1 shows a marginal increase ($+0.006$). A plausible explanation is that interpreting Japanese text within images—often vertical, multi-column, and annotation-heavy—remains difficult for current MLLMs, leading to unstable OCR and visual tokenization effects. Overall, the weighted average indicates that including images improves MC from 0.667 to 0.728 ($\Delta{=}+0.061$) and F1 from 0.429 to 0.479 ($\Delta{=}+0.050$), suggesting that multimodal input can improve performance on certain items, while also indicating remaining difficulties in visually grounded Japanese understanding.

\end{document}